\documentclass[letterpaper, 10 pt, conference]{ieeeconf}  % Comment this line out if you need a4paper

\IEEEoverridecommandlockouts                              % This command is only needed if 
\usepackage{graphics} % for pdf, bitmapped graphics files
\usepackage{epsfig} % for postscript graphics files
\usepackage{mathptmx} % assumes new font selection scheme installed
\usepackage{times} % assumes new font selection scheme installed
\usepackage{amsmath} % assumes amsmath package installed
\usepackage{amssymb}  % assumes amsmath package installed
\usepackage{bm}
\usepackage{algorithm}
\usepackage[noend]{algorithmic}
\usepackage{booktabs}
\usepackage{makecell}
\usepackage{stfloats}
\usepackage{cite}
\usepackage{titlesec}
\usepackage{color}
\usepackage{xspace}
\def \UAV{U\textsc{av}\xspace}
\def \UAVs{U\textsc{avs}\xspace}
\def \EANS{E\textsc{ans}\xspace}
\def \HIL{H\textsc{il}\xspace}
\def \ROS{R\textsc{os}\xspace}
\def \DL{D\textsc{l}\xspace}
\def \ASAP{A\textsc{sap}\xspace}

\title{\LARGE \bf
EANS: Reducing Energy Consumption for UAV with an Environmental Adaptive Navigation Strategy
}

\author{Tian Liu, Han Liu, Boyang Li*, Long Chen, Kai Huang% <-this % stops a space
\thanks{The authors are all with the school of computer science and engineering, Sun Yat-sen University, Guangzhou, China.}%
}

\begin{document}

\maketitle
% \thispagestyle{empty}
% \pagestyle{empty}

%%%%%%%%%%%%%%%%%%%%%%%%%%%%%%%%%%%%%%%%%%%%%%%%%%%%%%%%%%%%%%%%%%%%%%%%%%%%%%%%
\begin{abstract}
Unmanned Aerial Vehicles (\UAVs) are limited by the onboard energy. Refinement of the navigation strategy directly affects both the flight velocity and the trajectory based on the adjustment of key parameters in the \UAVs pipeline, thus reducing energy consumption. However, existing techniques tend to adopt static and conservative strategies in dynamic scenarios, leading to inefficient energy reduction. Dynamically adjusting the navigation strategy requires overcoming the challenges including the task pipeline interdependencies, the environmental-strategy correlations, and the selecting parameters. To solve the aforementioned problems, this paper proposes a method to dynamically adjust the navigation strategy of the \UAVs by analyzing its dynamic characteristics and the temporal characteristics of the autonomous navigation pipeline, thereby reducing  \UAVs energy consumption in response to environmental changes. We compare our method with the baseline through hardware-in-the-loop (\HIL) simulation and real-world experiments, showing our method 3.2X and 2.6X improvements in mission time, 2.4X and 1.6X improvements in energy, respectively.
\end{abstract}

%%%%%%%%%%%%%%%%%%%%%%%%%%%%%%%%%%%%%%%%%%%%%%%%%%%%%%%%%%%%%%%%%%%%%%%%%%%%%%%%
\section{INTRODUCTION}
Unmanned Aerial Vehicles (\UAVs) have been widely applied in a variety of fields\cite{saunders2024autonomous,paulin2024application,wu2021deep}. However, the limited onboard energy constraints the application of \UAVs. To solve the problem, numerous researches have been focused on reducing the energy consumption of \UAVs\cite{datsko2024energy, seewald2022energy, takemura2024perception}.

Refinement of the \UAVs navigation strategy is one of the methods to reduce energy consumption. The navigation strategy of \UAVs is the adjustment of parameters throughout the working pipeline \UAVs that includes perception, planning, and control, which directly affects both flight velocity and trajectory.  An efficient navigation strategy enhances the maneuverability of \UAVs during mission execution, resulting in faster flight velocity and shorter mission time\cite{boroujerdian2018mavbench}, thereby reducing the energy consumption.
%%Furthermore, research have indicated that reducing energy consumption can be achieved by adjusting navigation strategy, which encompass parameters related to perception, planning and control \cite{boroujerdian2021roborun}. Efficient navigation strategy can enhance the maneuverability of UAVs during mission execution, resulting in faster flight velocity and shorter mission time. Existing research confirms that UAVs can reduce energy consumption by reducing mission time\cite{boroujerdian2018mavbench}.

Existing researches \cite{zhou2019robust, zhou2020ego} commonly adopts the \textit{static} and \textit{conservative} strategy, leading to the constrained effectiveness in dynamic and complex environments. These strategies tend to overemphasize safety at the cost of flexibility and responsiveness particularly in scenarios with sparser obstacle distribution where the potential for collisions is inherently reduced. Therefore, there exists the energy reduction space by implementing a more aggressive navigation strategy, characterized by higher velocities and reduced perception precision. \textit{Thus, how to dynamically adjust the navigation strategy based on real-time environmental features to reduce energy consumption is the problem to be solved.}
% In the field of UAV navigation, current research \cite{zhou2019robust, zhou2020ego} commonly adopts \textit{static} and \textit{conservative} strategy, which prioritize mission success but are not necessarily optimal in every scenario. Specifically, in open environments, UAVs can leverage more aggressive navigation strategy, contrasting with the conservative strategy necessary for complex environments. Conservative strategy help minimize errors in state estimation, perception, and control, thereby reducing crash probabilities. In contrast, the lower obstacle density in open environments inherently reduces crash risks, enabling UAVs to adopt strategy that prioritize higher flight velocity and lower precision perception. Thus, \textit{dynamically adjusting} the navigation strategy can optimize mission time and energy consumption.

Unfortunately, dynamically adjusting the navigation strategy poses challenges. Firstly, the tasks within the navigation pipeline are highly interdependent and tightly coupled, indicating that the adjustment in one task affects the performance of others. Secondly, the multidimensional nature of the environmental information makes it difficult for the dynamic adjustment of navigation strategy in response to the environmental changes. Finally, with complicated \UAVs working pipeline, how to select and adjust the suitable parameters to reduce the energy consumption is still challenging. Existing research \cite{zhang2023asap,quan2021eva,zhao2024learning} focus on improving single aspects of the \UAVs pipeline, leading to the inadequate reduction in overall energy efficiency.
% However, trying to dynamically adjust the navigation strategy is difficult. Firstly, the components in the navigation pipeline are tightly coupled, and changing the parameters of one component can affect the results of another component. For example, perception results have a direct impact on planning decisions. Arbitrary changes to the perception parameters may lead to planning failure. Secondly,  as environmental information is multidimensional, adjusting navigation strategy in response to environmental changes is challenging. Finally, the navigation strategy contains many parameters, choosing which navigation parameters to minimize energy consumption is a challenge. Specifically, some of the research \cite{zhang2023asap,quan2021eva,zhao2024learning} has improved only a single component (e.g., perception or planning)  in the navigation pipeline, without adequately addressing the simultaneous adjustment of multiple components.

To address the aforementioned challenges, we propose the \textbf{E}nvironmental \textbf{A}daptive \textbf{N}avigation \textbf{S}trategy (\EANS) to reduce energy consumption for \UAVs. The corresponding advantages of the \EANS include: (i) The interdependencies among the tasks in pipeline are fully considered when dynamically adjusting the navigation strategy. (ii) The relationship between the environment and the navigation strategy is constructed based on the detailed analysis of the kinematic properties of \UAVs and the timing of navigation pipeline. (iii) The mapping frequency, the resolution, and the maximum velocity are all considered to reduce the energy consumption in \EANS. In addition, as a standalone component that communicates with the navigation pipeline, \EANS is designed to work with any probabilistic map-based navigation system, adaptable for various planners and controllers. To sum up, the contributions are as follows:

\begin{itemize} 
\item [1)]An environmental adaptive approach is proposed to dynamically adjust the navigation strategy to reduce the energy consumption of \UAVs.
\item [2)]The relationships between the environment and navigation strategy are analyzed by examining the kinematic characteristics and timing of navigation pipeline.
\item [3)]Extensive \HIL simulation and real-world experiments indicate the proposed \EANS efficiently reduces the energy consumption compared to baseline.
\end{itemize}

% The rest of the paper is organized as follows, related works are discussed in Sec.II. The problem formulation and system architecture are discussed in Sec.III. The main methods are discussed in Sec.IV and Sec.V. Experimental setup is described in Sec.VI and Sec.VII analyzes the experimental results. The paper is concluded in Sec.VIII.

\section{Related Work}
Adaptive navigation strategy for \UAVs has attracted extensive research interest. There have been a lot of work focused on adjusting the strategy in the planning component. In \cite{quan2021eva,wang2023speed}, researchers adapted the flight velocity in the planner by manually designing the cost function, which enables \UAVs to choose different flight strategies according to the complexity of the environment. However, manually designed cost functions are inherently less robust. Further, in \cite{richter2018bayesian,zhao2024learning,yu2024mavrl}, the researchers employed deep learning (\DL) to facilitate the autonomous learning of flight strategies in diverse environments. The implementation of \DL markedly enhanced the adaptability and flexibility of the strategy, enabling \UAVs to more effectively respond to environmental changes, achieving more precise and efficient navigation results. Although these approaches have seen success in adjusting planning strategy, they fail to fully leverage the low-precision perception strategy that \UAVs utilize in open environments, potentially missing the opportunities to further reduce energy consumption. Meanwhile, equipping \DL on \UAVs requires additional computing energy.

In addition to the adjustment of planning strategy, there are also researches on the perception strategy. \ASAP\cite{zhang2023asap} enables \UAVs to dynamically modulate perception frequency based on environmental complexity, thereby alleviating computational demands. However, \ASAP assesses complexity merely by counting sensor-detected obstacles using static thresholds. Further, combined with a lookup table, \ASAP exhibits parameter sensitivity and insufficient robustness across diverse environments. In contrast, the Roborun framework \cite{boroujerdian2021roborun} takes into account both the mapping resolution and the flight velocity of \UAVs to refining the strategy by evaluating module latency, demonstrating the notable performance improvements. Nevertheless, the Roborun framework is limited by the reliance on function fitting during the modeling process, which impacts its robustness. Moreover, this method does not account for the adjustment of the frequency, which also affects the energy consumption of \UAVs.

\section{Background Knowledge}
The energy consumed by \UAVs in navigation missions primarily consists of both flight and computing energy\cite{sudhakar2020balancing}.  We reduce the energy $E$ by adjusting the \UAVs flight velocity $v_t$, the mapping frequency $H$, and the resolution $R$ in \UAVs pipeline, described in the following equation:
\begin{equation}
\label{equ1}
\mathop{\arg\min}\limits_{v_t,H,R}\  E=\int_0^{t_f} g(v_t) \ dt +\int_0^{t_c}f(u_t) \ dt 
\end{equation}
where $t_f$ and $t_c$ represent the total time consumed by \UAVs during flight and computation. $g(v_t)$ and $f(u_t)$ describe the flight and computational power consumption of \UAVs at timestamp \textit{$t$}. The computational power consumption is affected by a number of factors $u_t$, including but not limited to mapping frequency, resolution, and planning search time. In order to effectively reduce the total energy consumption of \UAVs, there are three possible improvement directions: 
\begin{itemize}
\item [a)]Reducing flight time $t_f$ or computation time $t_c$. By increasing the average flight velocity of \UAVs for a given navigation distance, the required time to the mission is reduced, thus reducing energy consumption\cite{boroujerdian2018mavbench}.
\item [b)] Reducing the computational power consumption $f(u_t)$. For micro \UAVs, their computing power is similar to flight power. Computing for 1 second is as expensive as flying another 15.3 meters\cite{sudhakar2020balancing}. Therefore, it is particularly important to save computing power through efficient computational strategies.
\item  [c)] Adopting more advanced flight control strategy to reduce the flight power consumption $g(v_t)$. This involves precise control of flight velocity and the acceleration to reduce energy consumption.
\end{itemize}

Therefore, based on directions a) and b), we choose to adjust the flight velocity, the mapping frequency and the resolution in navigation strategy to reduce energy. Specifically, flight velocity affects mission duration and thereby influencing flight energy consumption. Perception accuracy impacts the amount of onboard computation, which affects computational energy consumption. Consequently, we selected these parameters as the environmental adaptive navigation strategy.

\section{Problem Formulation}
Directly solving the Eq. \eqref{equ1} is difficult, since there are numerous variables and abstract functions involved. Consequently, we decompose the problem into two distinct subproblems: one focused on flight velocity and the other on mapping parameters.

\begin{figure}[t]
      \centering
      \includegraphics[width=.9\linewidth]{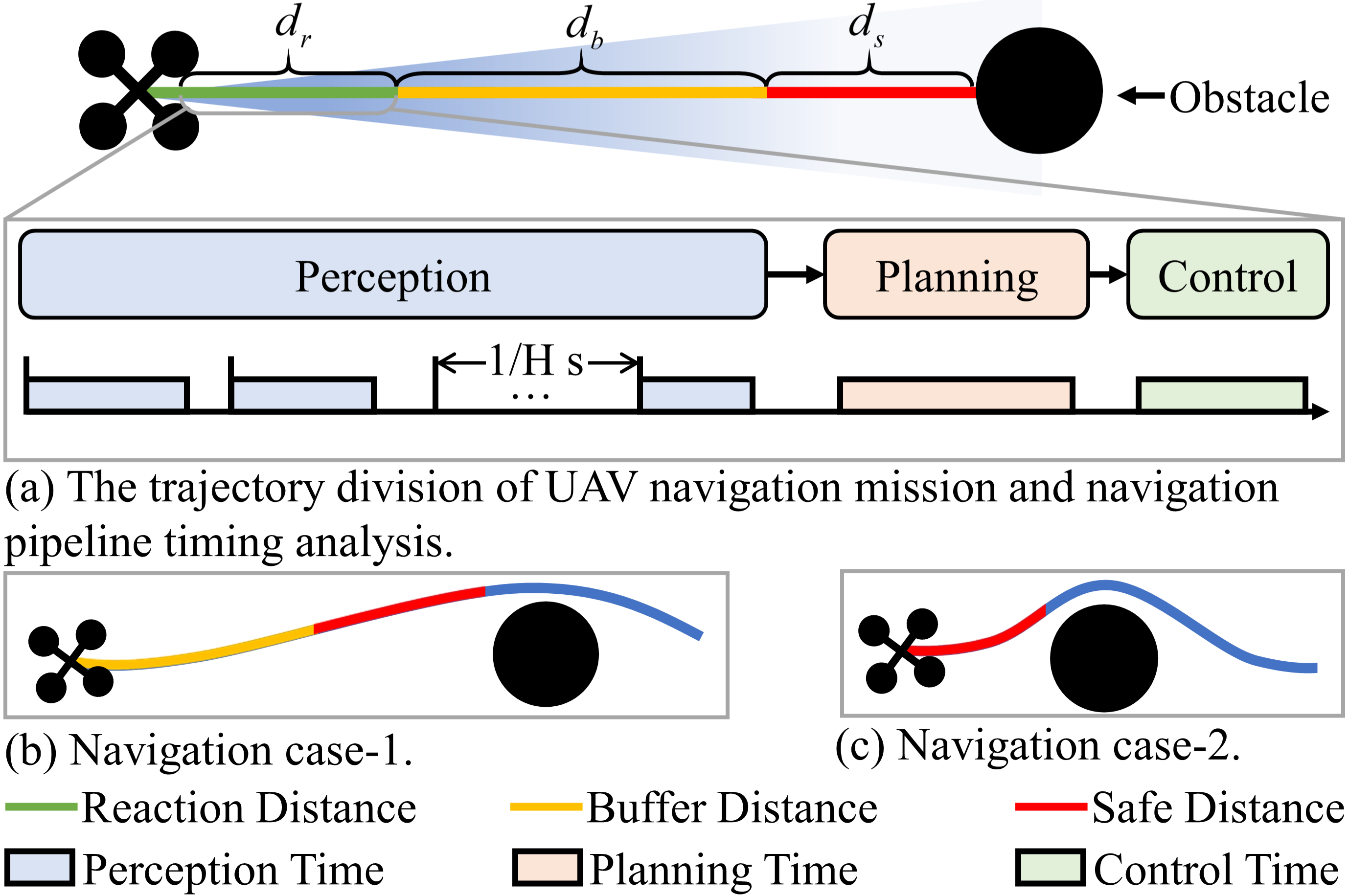}
      \caption{Division of trajectory, timing analysis of pipeline, and contrast of two navigation cases at different flight velocities.}
      \label{traj_division}
    \vspace{-10pt}
   \end{figure}

For an obstacle with distance $d$, \UAVs need to perform an obstacle avoidance mission. Therefore, the existing trajectory is divided into three parts. The first part is the reaction distance $d_r=v_tt_r$, where $v_t$ is the flight velocity of \UAVs, $t_r$ is the time interval required for the \UAVs to detect an obstacle and initiate a response. The second part is the safe distance $d_s$. \UAVs need to reserve enough distance to execute the response. Considering the worst case, where \UAVs need to make an emergency stop, $d_s$ is defined as follows:
\begin{equation}
\label{equ2}
d_s=\frac{v_t^2}{2a_{max}}
\end{equation}
where $a_{max}$ is the maximum acceleration of \UAVs. The third part is the buffer distance $d_b=d-d_r-d_s$. We noticed that $d_b$ is related to $v_t$. Given that $d$ remains constant, an increase in $v_t$ leads to the increase in both $d_s$ and $d_r$, thereby diminishing $d_b$. This indicates that the obstacle avoidance distance of the \UAVs is shortened when the velocity of the \UAVs increased. Conversely, by effectively managing the buffer distance, the velocity of the \UAVs increase while maintaining the successful obstacle avoidance.

The division of the trajectory and the timing of the navigation pipeline are shown in Fig. \ref{traj_division}(a). According to the principle of grid map updating \cite{elfes2013occupancy}, the sensor needs repeatedly identify an unknown obstacle within a grid several times to transition the grid state from unknown to occupied. Consequently, the perception process is executed multiple times. Fig. \ref{traj_division}(b) and Fig. \ref{traj_division}(c) show the obstacle avoidance at slower and faster velocities of the \UAV, respectively. In Fig. \ref{traj_division}(b), the \UAV flies slowly, thus having a more generous obstacle avoidance distance. In Fig. \ref{traj_division}(c), despite having a reduced obstacle avoidance distance, the \UAV exhibits a higher velocity. Therefore, to increase the velocity of \UAVs,  Subproblem 1 is summarized as follows:
\begin{equation}
\label{subproblem1}
\mathop{\arg\min}\limits_{v_t}\ d-d_r-d_s
\end{equation}

In addition, we shorten $t_c$ and $f(u_t)$ by adjusting the mapping frequency $H$ and the resolution $R$ to further reduce energy efficiency. Thus Subproblem 2 is as follows:
\begin{equation}
\label{subproblem2}
\mathop{\arg\min}\limits_{H,R}\  \int_0^{t_c}f(H,R) \ dt 
\end{equation}

In fact, establishing a direct mathematical correlation between $H$, $R$, $t_c$, and $f(u_t)$ is challenging. However, previous studies have shown that lowering $H$ and $R$ can reduce $t_c$ and $f(u_t)$\cite{zhang2023asap, boroujerdian2021roborun}. Accordingly, we aim to minimize the mapping frequency and resolution, taking into account the relevant constraints.

\begin{figure}[t]
  \centering
  \includegraphics[width=1\linewidth]{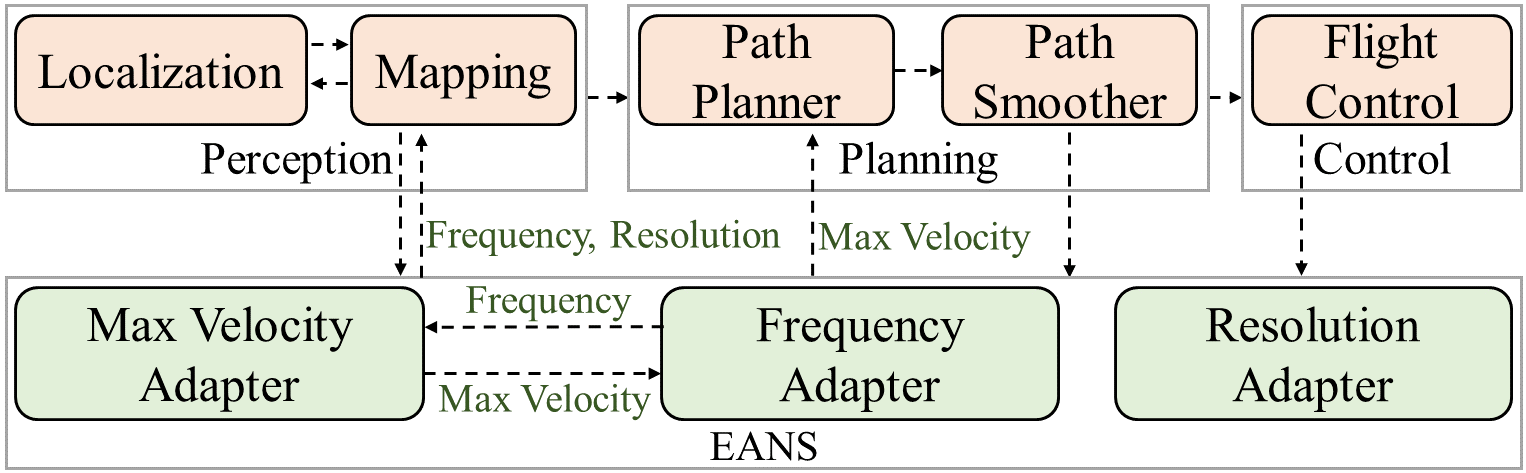}
  \caption{System architecture, navigation pipeline (red), \EANS including three adapters (green). Arrows show the key labelled dataflow.}
  \label{architecture}
\vspace{-10pt}
\end{figure}

In \EANS, we designed three adapters to solve the aforementioned two subproblems. As shown in Fig. \ref{architecture}, the system architecture is divided into two main parts: the red squares show the navigation pipeline of \UAVs. The green parts show the \EANS. The three adapters calculate the most appropriate navigation strategy based on the obtained dataflow and pass strategy back to the navigation pipeline. 

\section{Environmental Adaptive Navigation Strategy}
This section details the design of three adapters for adjusting the navigation strategy in \EANS.
\subsection{Maximum Velocity Adapter}

Solving Eq. \eqref{subproblem1} by replacing $d_r$ and $d_s$\cite{boroujerdian2018mavbench}:

\begin{equation}
    \label{equ3}
    v_{t} \leq \sqrt{(t_ra_{max})^2+2da_{max}}-t_ra_{max}
\end{equation}

Obviously, the upper limit of flight velocity \( v_t \) of the \UAVs is affected by the distance \( d \) and the response time \( t_r \). 

% \subsection{Effective Perceived  Distance}
To simplify the expression, we define the distance required by the \UAVs sensor to capture an obstacle as the effective perceived distance \( d \), which is mainly limited by the maximum detection range \( d_c \) of the sensor and the environment. When flying in open environments, \( d \) is equal to \( d_c \). However, in complex and unknown environments, due to factors such as visual occlusion and motion blurring, there exists the situations where \( d \) is less than the \( d_c \). Therefore, an accurate description of \( d \) for different environmental conditions is essential for determining the \( v_t \) of the \UAVs.

Inspired by \cite{quan2021eva}, we propose a method to consider the nearest distance between the planning trajectory and the obstacle, as well as the angle between velocity direction and the gradient. In order to accurately describe \( d \), the closest distance between the trajectory and an obstacle is determined by the two endpoints \( \bm{P}_t \) and \( \bm{P}_o \), defining as the gradient \( \nabla c \).
\begin{equation}
    \label{equ4}
    \nabla c=-\frac{\bm{P}_o-\bm{P}_t}{\Vert\bm{P}_o-\bm{P}_t\Vert_2 }
\end{equation}

We use a sigmoid function to map the cosine of the angle between the velocity vector of the \UAVs as $\bm{V}$ and the gradient $\nabla c$ to the risk weight $\eta \in [0,1]$, with the function defined as: 
\begin{equation}
    \label{equ5}
    \beta = \frac{<\bm{V},\nabla c>}{\Vert\bm{V}\Vert\Vert\nabla c\Vert}
\end{equation}
\begin{equation}
    \label{equ6}
    \eta = \frac{1}{1+e^{\alpha\beta}}
\end{equation}
where $\alpha \in \mathbb{R}^+$ is the rate of changing coefficient. $<,>$ is the dot product operator. The closest distance from the trajectory to the obstacle is $\Vert \nabla c \Vert$. Therefore, the effective perceived distance $d$ is described as:
\begin{equation}
    \label{equ7}
    d=\lambda(1-\eta)d_c+(1-\lambda)\Vert \nabla c \Vert
\end{equation}
where $\lambda \in (0, 1)$ balances the influence of \(\eta\) , representing the angle between the velocity direction and the gradient. \( \lVert \nabla c \rVert \) represents the distance to the nearest obstacle. In particular, the maximum value of \( \lVert \nabla c \rVert \) in an open environment is \( d_c \). Furthermore, if no obstacles are detected within the  sensing range, the risk weight \( \eta \)  equals zero. By incorporating these conditions into Eq.\eqref{equ7}, we obtain $d=d_c$, which aligns with the previous assumption.

% \begin{figure}[t]
%   \centering
%   \includegraphics[width=1\linewidth]{img/2.png}
%   \caption{Mapping, planning and others pipeline thread time schematic}
%   \label{fig_2}
% \end{figure}
% \section{Environmental Adaptive Perception}
% This section will further elaborate on the principles of operation of the mapping frequency and resolution adapters.
\subsection{Mapping Frequency Adapter}
The time $t_r$ required by \UAVs to complete a navigation pipeline consists of three main components: (i) The time $t_m$ required from the capture of the obstacle by the sensor to the determination of the obstacle by the mapping algorithm. (ii) The time $t_p$ required to perform a single motion planning. (iii) Others time $t_o$, including the conversion of the transmitted command. Setting the number of captures performed by the sensor on the obstacle as \( \sigma \), the relationship between \( t_m \) and the single run time of the mapping algorithm \( \hat{t}_m \) can be expressed as follows:
\begin{equation}
    \label{equ8}
    t_m=\frac{\sigma-1}{H}+\hat{t}_m
\end{equation}
The condition \( \hat{t}_m \leq \frac{1}{H} \) must be met, meaning that the time taken for one diagram update should not exceed the interval between updates, which has been demonstrated in the navigation timing analysis in Fig. \ref{traj_division}(a). Incorporating \( t_m \) into the calculation of \( t_r \) and combining it with Eq. \eqref{equ3}:
\begin{equation}
    \label{equ9}
    \begin{aligned}
        &\hat{t}_s = \hat{t}_m+t_p+t_o&\\
        &\frac{2(\sigma-1)a_{max}v_t}{2da_{max}-v_t^2-2a_{max}v_t\hat{t}_s} \leq H \leq \frac{1}{\hat{t}_m}&
    \end{aligned}
\end{equation}
The mapping frequency $H$ is mainly limited by the flight velocity $v_t$, the effective perceived distance $d$, and the computation time $\hat{t}_s$. Experimental data show that reducing the mapping resolution can reduce $\hat{t}_m$\cite{boroujerdian2021roborun}. However, low precision mapping resolution may adversely affect the navigation performance of \UAVs in complex environments. Therefore, dynamically adjusting the mapping resolution according to environmental variations is crucial for \UAVs to accomplish their mission efficiency.

\begin{algorithm}[!t]
    \caption{Mapping Resolution Adapter Algorithm}
    \label{algo_1} 
    \begin{algorithmic}[1]
        \REQUIRE Trajectory $L$, Resolution list $R$, Threshold $\phi$
        \ENSURE Resolution $r_c$

        \STATE Update $P$ based on Equation \eqref{equ11}
        \STATE $r_c \gets R_0$
        \IF {$P \geq \phi$}
             \FOR{$r=R_1$ to $R_{n}$ }
                \IF {manhattanDistance($\bm{P}_t$, $\bm{P}_o$, $r$) $>0$}
                    \STATE $r_c \gets r$
                \ELSE
                    \STATE \textbf{break}
                \ENDIF
            \ENDFOR
        \ENDIF
    \RETURN $r_c$
	\end{algorithmic} 
\end{algorithm}

\begin{figure}[t]
  \centering
  \includegraphics[width=1\linewidth]{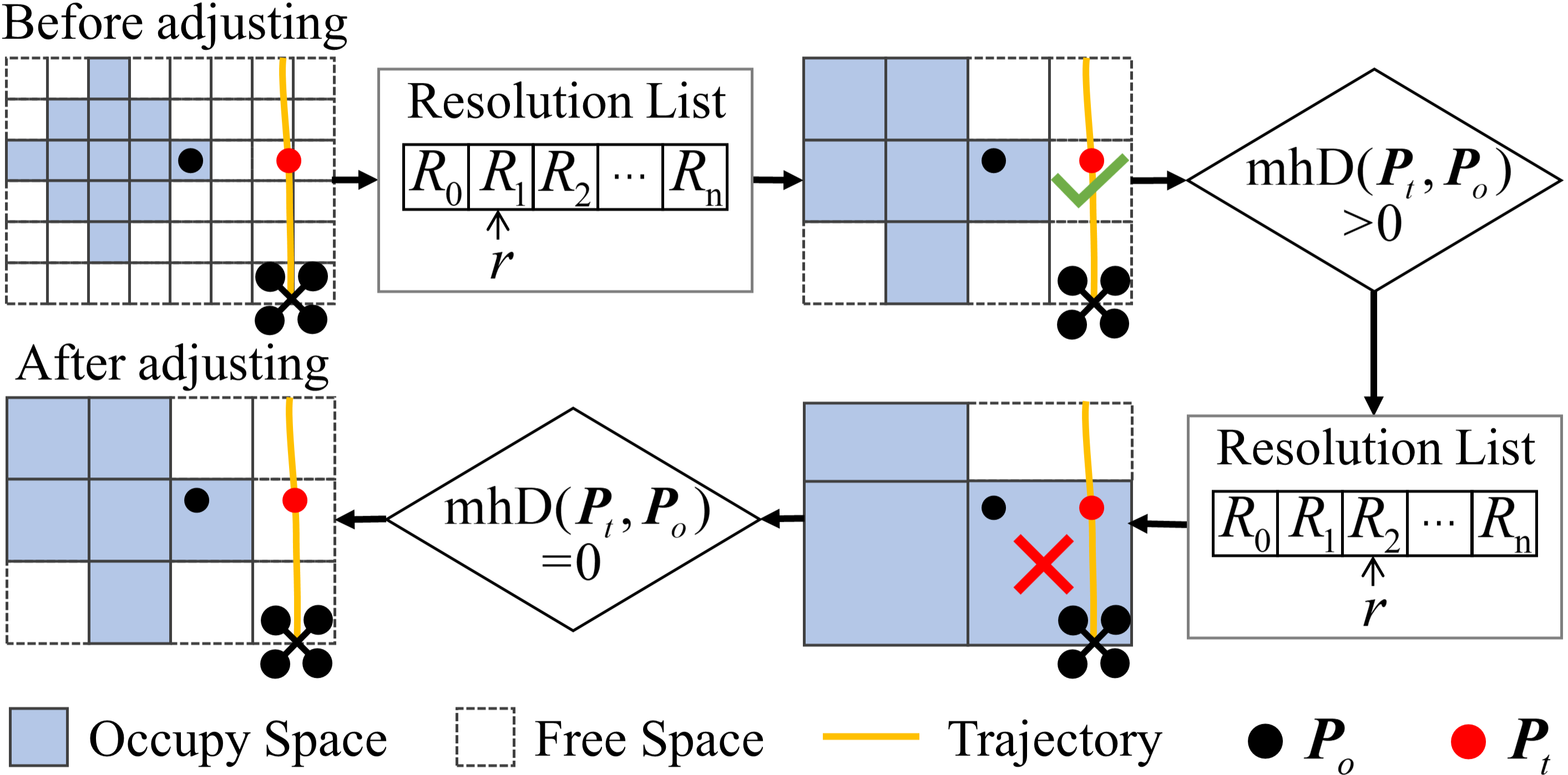}
  \caption{Process example of the mapping resolution adapter. The mhD is the Manhattan distance computation function.}
  \label{fig_4}
  \vspace{-10pt}
\end{figure}
\subsection{Mapping Resolution Adapter}
The mapping resolution adapter process is illustrated in Alg. \ref{algo_1}, which includes two key steps. %It helps to reduce the effects of resolution changes on current routes and avoids disrupting the planning process.  

\textbf{Computing better trajectory probability}: The adapter refines mapping resolution based on better trajectory probability. If probability exceeds a preset threshold \(\phi \), the adapter will try to reduce the mapping resolution. In contrast, the resolution will increase to avoid unnecessary trajectory detours. The length of the planned trajectory from the current position of the \UAVs to the target position is \( L \). The euclidean distance between two endpoints positions is \( \bar{L} \). An optimal trajectory length \( L^* \) is calculated as follows:
\begin{equation}
    \label{equ10}
    L^*=(1+\gamma\frac{\rho_c}{\rho^*_c})\bar{L} 
\end{equation}
where $\rho_c$ and $\rho^*_c$ denote the number and the maximum number of obstacle pixels captured by the sensor. $\gamma\in\mathbb{R}^+$ is the scale factor. The optimal trajectory length \( L^* \) should approach the direct euclidean distance \( \bar{L} \) and increase with higher obstacle density in the scenario. We define the probability \( P\{T=b\} \) of the current trajectory being suitable based on these conditions:
\begin{equation}
    \label{equ11}
    P\{T=b\}=e^{-\frac{\epsilon}{\bar{L}}{(L-L^*)}}, L\geq L^*
\end{equation}

To ensure the current trajectory is suitable, we require \( L \geq L^* \). Evidently, with the increase of Euclidean distance $\bar{L}$, the influence of fixed length error on trajectory detour probability will decrease. Therefore \( \epsilon / \bar{L} \) measures how a fixed length error affects the trajectory probability at varying distances.

\textbf{Adjust mapping resolution}: If a lower mapping resolution is required, the adapter uses a greedy algorithm to find the map index values at the new resolution for the endpoints \( \bm{P}_t \) and \( \bm{P}_o \). The process continues until the lowest resolution is reached, at which point the Manhattan distance between the indices of the two endpoints is nonzero. This ensures that the grids associated with the endpoints are in proximity but do not overlap. The adjusted map does not interfere with the current trajectory.

If the better trajectory probability is less than \(\phi\), the adapter tunes the mapping resolution to the highest value and perform to replanning the trajectory. Fig. \ref{fig_4} shows an example of adjusting the map resolution, where black and red dots represent endpoints $\bm{P}_o$ and $\bm{P}_t$, respectively. When the resolution is adjusted to $r=R_1$, obstacles expand due to the decreased resolution. However, the Manhattan distance between the two endpoints remains greater than 0, thus allowing the adoption of this resolution. Upon further reducing the resolution, the Manhattan distance between the two endpoints drops to 0, rendering the resolution unacceptable. Consequently, the process terminates, and the previously suitable resolution is retained.

\section{Experiments}
This section provides a detailed description of the experimental setup and the evaluation results of the hardware-in-the-loop simulation and the real-world experiments.
\subsection{Experimental Setup}
Compared to complete simulation experiments, \HIL simulation experiments can more effectively verify the performance of algorithms on onboard computing platforms, providing essential preliminary validation for real-world experiments. In the hardware-in-the-loop simulation experiments, we use the Robot Operating System (\ROS) and the Gazebo to build the simulation platform. The environment simulation is powered by a high-performance PC with an AMD Ryzen5 3600 CPU and NVIDIA 1660 GPU. The \UAV's autonomous navigation is equipped on an Nvidia Jetson TX2 board. 

In addition, we craft numerous scenes, including parks, caves, and other environments to simulate real-world flight conditions. When designing simulation scenes, we ensure that the environment has sufficiently varied and realistic changes. For example, in the park environment, there are complex forest and architectural scenes as well as completely open areas. These open areas may be due to the presence of streets or rivers, which are common and realistic features in real-world scenarios. We track performance metrics including computation time, trajectory length and flight velocity for evaluation. The energy consumption for computation and flight during navigation is quantified. Computational energy is directly measured, whereas flight energy consumption is estimated based on the method referred in \cite{yu2023avoidbench}.

\begin{table}[!t]
  \caption{Adapters setting for three designs.\label{tab:table2}}
  \vspace{-10pt}
  \begin{center}
  \resizebox{\linewidth}{!}{
    \begin{tabular}{cccc} \hline 
       Variables& Baseline &Lookup Table& Proposed\\
        \midrule
        Velocity (m/s)& 0.5&[0.5-2.5]& [0.5-3.5]\\ 
 Frequency (fps)& 30 &[5, 10, ..., 30]& [5, 10, ..., 30]\\
 Resolution (m)& 0.1&[0.1,0.15, ...,0.5]&[0.1, 0.15, ..., 0.5]\\\hline
    \end{tabular}
  }
  \end{center}
  % \vspace{-15pt}
\end{table}

We choose the model-based trajectory planner, EGO-planner\cite{zhou2020ego} as the baseline for static navigation. The parameters for this baseline are set based on the settings in \cite{boroujerdian2021roborun} and \cite{FastDrone250}. In addition, we implemented the Lookup Table\cite{zhang2023asap} method for comparing the advantages of our method. All the specific parameter settings are displayed in Tab. \ref{tab:table2}.

Ultimately, our method is implemented on a quadrotor \UAV and evaluated against the baseline in navigation missions within cluttered environments.

\subsection{Experimental Results}
This section demonstrates the performance of the three designs across various environments, comparing overall results, analyzing the performance of the methods on a representative mission, and assessing their effectiveness in environments with varying levels of difficulty. Finally, the proposed method is validated in real-world experiments.

\begin{table}[!t]
  \caption{Mission level metrics for three designs.\label{tab:table3}}
  \vspace{-5pt}
  \begin{center}
  \resizebox{\linewidth}{!}{
    \begin{tabular}{cccc} \hline 
    & \makecell{Mapping Total \\ Time (ms)}& \makecell{Mapping Avg \\ Time (ms)}& \makecell{CPU Avg \\ Utilization (\%)}\\
        \midrule
        Baseline& 54792.44& 23.56& 79.11\\
 Lookup Table& \textbf{7152.38}& \textbf{11.60}&\textbf{43.90}\\ 
 Proposed& 9344.54& 14.15& 55.90\\
        \midrule
 & \makecell{Trajectory  Length (m)}& \makecell{Flight Time (s)}& \makecell{ Energy (\%)}\\
        \midrule
 Baseline& \textbf{59.63}& 117.52& 100\\
 Lookup Table& 64.50& 54.93&63.45\\
 Proposed& 60.32& \textbf{36.38}& \textbf{42.27}\\\hline
    \end{tabular}
  }
    \end{center}
\vspace{-15pt}
\end{table}

\subsubsection{Overall Results}
We conducted three distinct designs across seven simulated environments, with each design undergoing a minimum of five replicate experiments within each environment. The average metrics for all results are presented in Tab. \ref{tab:table3}, where the Mapping Avg Time is the average single time spent in the mapping component, the Mapping Total Time represents the total time spent in the mapping component during the completion of the mission.

Comparing the baseline, the Lookup Table and the proposed method improve the flight time by 2.1X and 3.2X, respectively by allowing the \UAV to fly at higher velocity in an open environment. Since the Lookup table lacks integration of mapping frequency and flight velocity, it decreases velocity to accomplish mission successfully. Both dynamic adjustment methods lower resolution, increase flight distance. However, the Lookup Table method exhibits the increase in trajectory length compared to the proposed method, attributed to its fixed threshold and the inability to consider the impact of trajectory detours.

To facilitate comparison, energy consumption is normalized and presented as a percentage, with the baseline energy consumption set to a benchmark value of 100\%. The Lookup Table and the proposed method improve energy consumption by 1.6X and 2.4X, respectively, compared to the baseline, due to the faster completion of the navigation mission and lower computational consumption of the two dynamic methods. The Lookup Table method reduces less energy than proposed method because of its maximum flight velocity limitation and increased trajectory length.

\begin{figure}[t]
  \centering
  \includegraphics[width=1\linewidth]{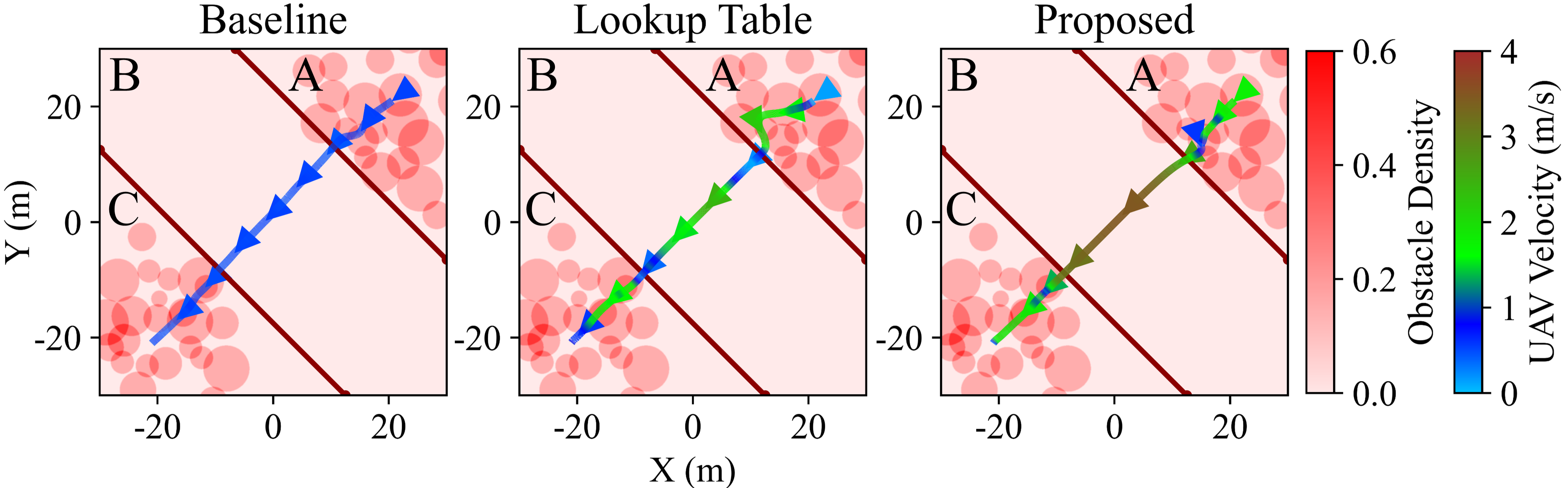}
  \caption{Analysis of representative missions. The heat map shows the distribution of obstacles, and the curves and their colors indicate the \UAV's flight trajectory and flight velocity.}
  \label{fig_5}
\end{figure}
\begin{figure}[t]
  \centering
  \includegraphics[width=1\linewidth]{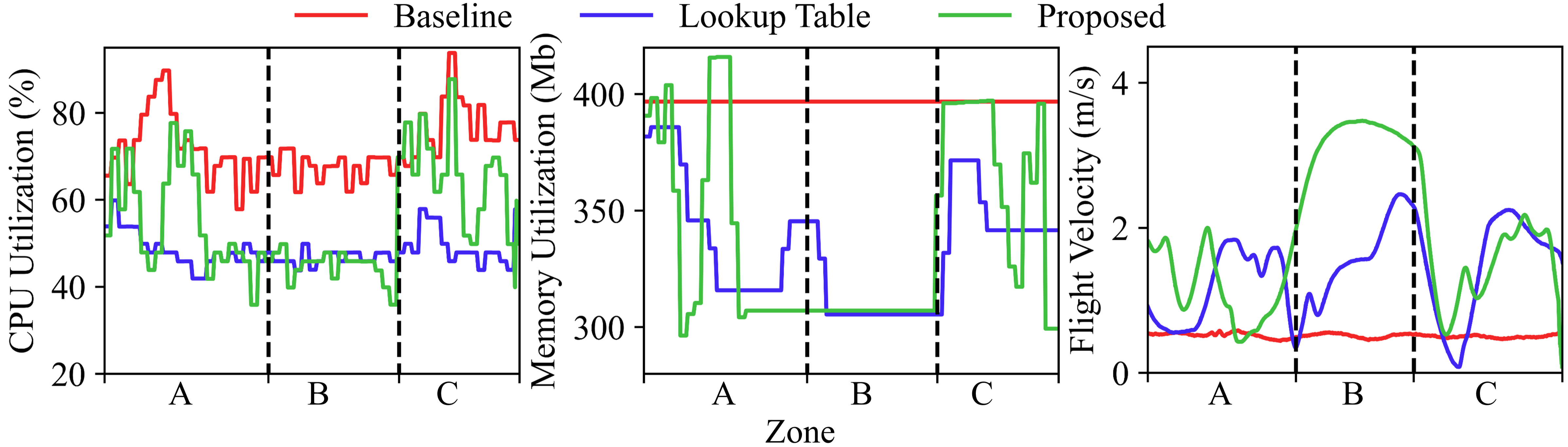}
  \caption{CPU, memory utilization, and flight velocity curves for three designs in representative missions. Zones A, B, and C are the areas in Fig. \ref{fig_5}. The red, blue, and green curves represent the baseline, Lookup Table, and proposed method, respectively}
  \label{fig_6}
\vspace{-10pt}
\end{figure}

The proposed method and the Lookup Table significantly reduce the mapping time and CPU usage. Compared to baseline, the Lookup Table improves the total mapping time and the average mapping time by 7.7X and 2.0X, improving the CPU utilization by 44.5\%. The proposed method improves the above metrics by 5.9X, 1.7X and 29.3\%, respectively. These metrics of the proposed method is slightly higher than the Lookup Table, as it tries to search for trajectory in complex environments using the higher mapping frequency and the corresponding resolution.

% \vspace{5pt}

\subsubsection{Representative Mission Analysis}

Fig. \ref{fig_5} shows the correlation between obstacle density and flight velocity. Within the region of the X-coordinate approaching to -10 m, both dynamic methods are decreasing the flight velocity to deal with the challenges of complex environments. As the X-coordinate approaches to 0 m where an open environment with sparse obstacle, the proposed method is able to utilize the computational resources more efficiently and improve the efficiency of the mission execution. Conversely, the baseline approach maintains a consistently lower flight velocity, attributed to its inherent static characteristics. Furthermore, the Lookup Table method fails to account for the influence of adjusting mapping resolutions on planning tasks. It tends to produce trajectory detours in regions where X-coordinate approaches to 20 m.

\begin{figure}[t]
  \centering
  \includegraphics[width=1\linewidth]{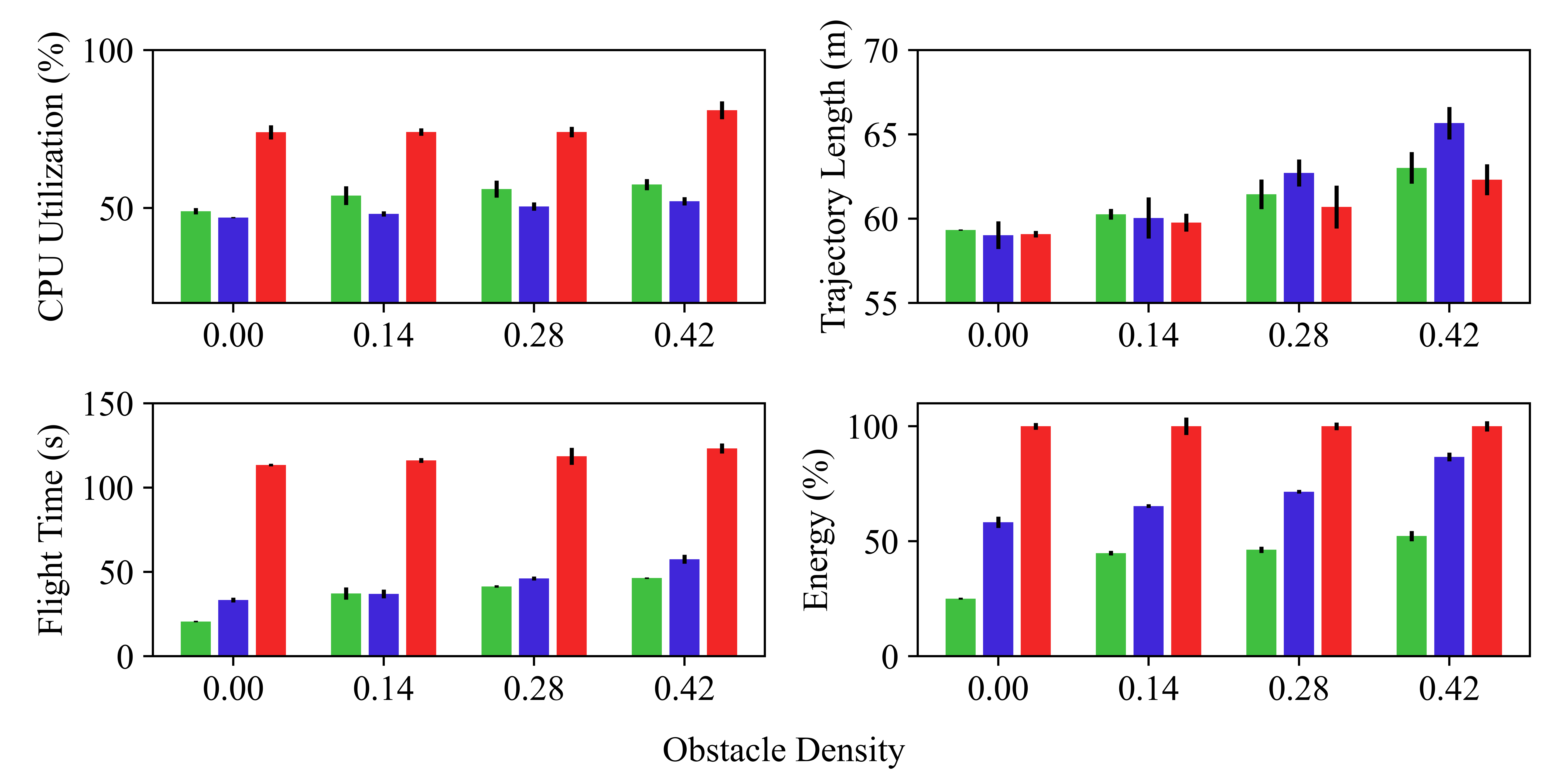}
  \caption{Comparison of performance metrics of three designs under different environmental difficulties. The abscissa is the obstacle density, and the higher the obstacle density, the more obstacles in the area, and the higher the difficulty for the \UAV to cross.}
  \label{fig_7}
% \vspace{-10pt}
\end{figure}

% \vspace{5pt}

\begin{figure*}[b]
      \centering
      \includegraphics[width=1\linewidth]{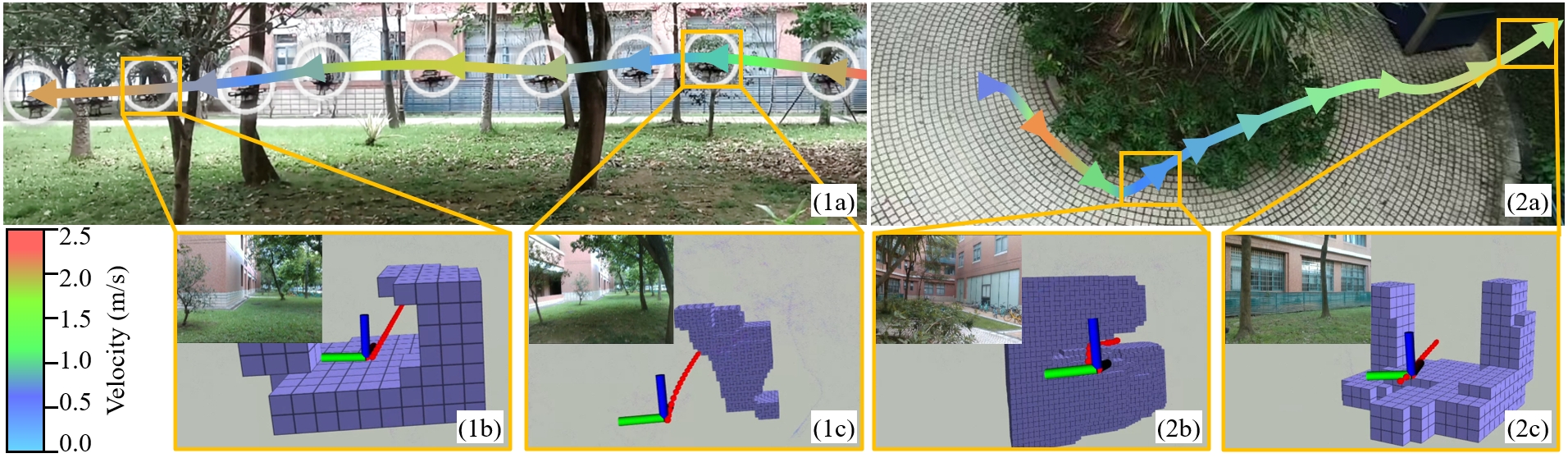}
      \caption{The \UAV flies in two environments. Subfigures (a), (b) and (c) respectively show the trajectory of the \UAV with flight velocity and different navigation strategies of the \UAV in different scenarios.}
      \label{fig_8}
% \vspace{-10pt}
\end{figure*}

For the three zones (Zone A, B, C) divided in Fig. \ref{fig_5}, Fig. \ref{fig_6} compares the computational resource consumption and flight velocity in different zones in detail. In the open environment (Zone B), the dynamic adjustment method not only improves the flight velocity, but also improves the use of computational resources, showing higher resource utilization efficiency and faster mission completion. In environment with dense obstacles (Zone A, C), the proposed method prevents collisions by reducing the flight velocity of the \UAV. In addition, the computational resource consumption of the proposed method is closed to the baseline with the same perception precision, demonstrating the lightness of the proposed method.

The Lookup Table fails to fully utilize the high resolution mapping to explore better trajectories in Zone A and C due to its fixed threshold limitations. Although the method reduces the use of computational resources, its maximum flight velocity is limited, leading to several emergency breaking, which affect the continuity and efficiency of the mission.

\subsubsection{Comparison of Environmental Difficulty}
In Fig. \ref{fig_7}, the proposed method exhibits dynamic adjustment capabilities in response to environmental challenges across a spectrum of complexities. In contrast, the baseline method is constrained by the static parameters, showing limited adaptability with metrics remaining largely constant. Despite aligning with the performance trend of the proposed method, Lookup Table exhibits limitations in flight time and trajectory length due to its fixed threshold constraint.  

In open environments, the proposed method improves computational energy efficiency while utilizing fewer computational resources. As obstacle density increases, the method's resource allocation for perception increases, improving adaptability in complex environment. Although it shows a slight decrease in trajectory length compared to the baseline, the proposed method outperforms in other metrics due to its greedy strategy, which improves resource allocation. The Lookup Table method's fixed threshold makes it less responsive to detours. Consequently, it performs poorly in flight time and energy efficiency compared to the proposed method, despite requiring fewer computational resources.

% \vspace{5pt}

\subsubsection{Real-World Experiments}
We selected two experimental environments under real-world conditions: a typical cluttered area (Fig. \ref{fig_8}(1a)) and a narrow slit environment (Fig. \ref{fig_8}(2b)) with increased complexity. The algorithm is deployed on a quadrotor with the size of 55 cm × 55 cm × 35 cm and evaluated in two different real-world environments. The \UAV's maximum flight velocity is set to 2 m/s. The perception strategy aligns with the simulation experiment configuration. Fig. \ref{fig_8} shows the \UAV flying trajectory in the real environment. Environment (1) represents a woodland scenario in \UAV navigation mission. Environment (2) has large and irregular obstacles, necessitating advanced \UAV perception capabilities. 

  \begin{figure}[t]
  \centering
  \includegraphics[width=1\linewidth]{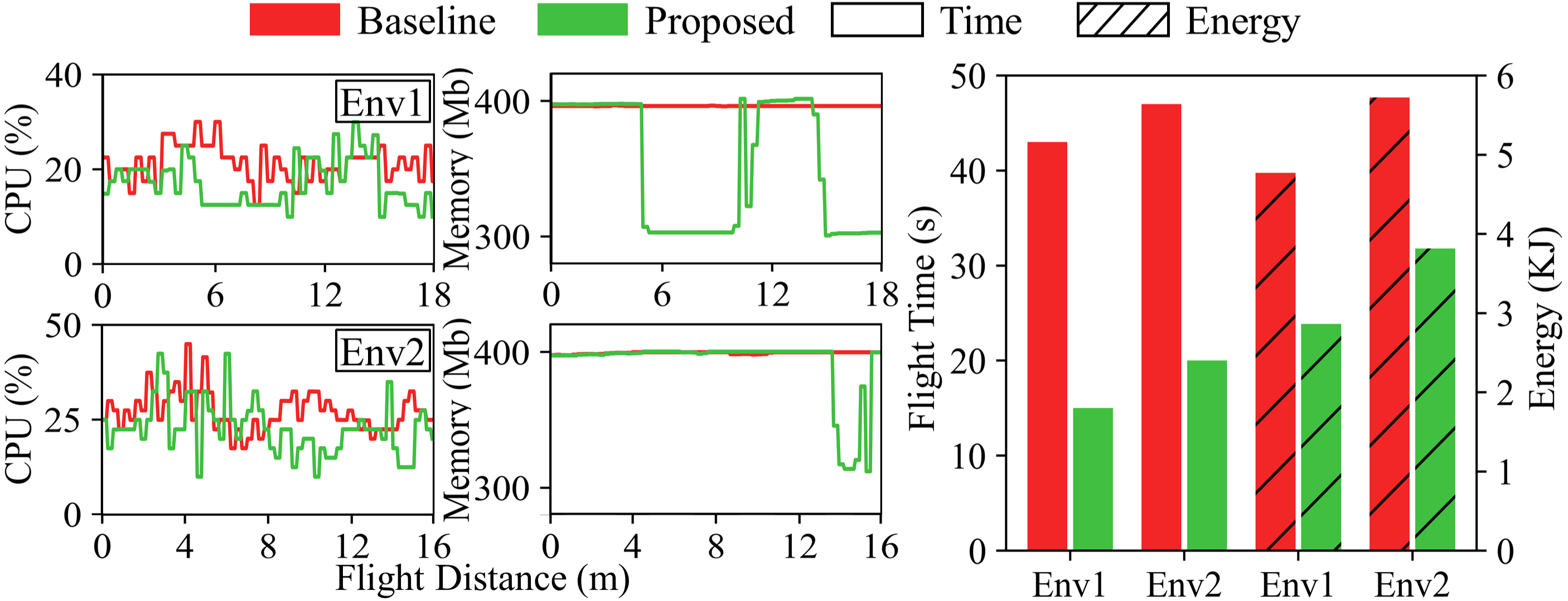}
   \caption{Computational resources utilization curves and flight time, energy consumption data for the baseline and proposed method in real-world experiments. }
  \label{fig_9}
% \vspace{-10pt}
\end{figure}

In Environment (1), the \UAV adjusts velocity, decelerating near trees and accelerating in open areas, to reduce energy consumption. Fig. \ref{fig_8}(1b) and (1c) show the aggressive and conservative perception strategies. Specifically, in (1b), a higher resolution map is used for obstacle evasion, reducing trajectory deviations. In (1c), a lower resolution map is utilized for distant obstacles, reducing computational costs.

In Environment (2), the \UAV maintains high precision resolution in the early stage (Fig. \ref{fig_8}(2b)) of navigation mission to ensure successful passage through narrow spaces. As the \UAV flies to more open areas (Fig. \ref{fig_8}(2c)), the resolution decreases accordingly. This strategy effectively enhances the resource allocation during the navigation mission, thereby improving navigation efficiency. 

The results regarding \UAV flights in two distinct environments, including computational resource utilization, flight duration, and total energy consumption, are shown in Fig. \ref{fig_9}.  The dynamic adjustment of the perception strategy results in improvements over the baseline in Environment (1), with an average reduction of 18.9\% in CPU utilization and 11.4\% in memory utilization. In Environment (2), the \UAV achieves a CPU utilization reduction of 15.6\% and a memory utilization reduction of 2.0\%. The reduced memory savings in Environment (2) relative to Environment (1) are attributed to the \UAV's necessity to maintain high precision perception during the early phase of navigation mission, which is essential for improving navigation efficiency in complex environments.

In addition, in two different real-world environments, the \UAV utilizing the baseline algorithm exhibites an average mission completion time of 45 s with energy consumption of 5.25 kJ. Comparatively, the \UAV implementing the proposed method achieved a reduction in average mission time to 17.5 s with energy consumption to 3.34 kJ. The experimental results indicate that the proposed method results in 2.6X improvement in mission time and 1.6X improvement in energy consumption, thereby substantiating the enhanced performance of the proposed algorithm relative to the baseline parameters.

% In the experiments for the navigation mission with a length of about 18m, the UAV took about 43s to complete the navigation mission with an energy consumption of 4.77KJ when using a navigation strategy that was consistent with the parameters of the simulated baseline. In contrast, after applying the proposed method, the time taken by the UAV to complete the same navigation mission is reduced to about 15s and the energy consumption is reduced to 2.862KJ. The experimental results show that the proposed method achieves a reduction of 65.1\% in mission time and 40\% in energy consumption, thus effectively validating the performance improvement of the algorithm compared to the baseline parameters. 

\section{CONCLUSIONS}

In this paper, we propose an environmental adaptive navigation strategy to reduce energy consumption by analyzing the kinematic properties of the \UAVs and the timing of the navigation pipeline. Compared with the baseline and other adjustment methods, our method completes the navigation mission more quickly, effectively reducing flight and computational energy consumption. In the future, we plan to further explore the impact of our algorithm under different flight and computational energy ratios, and determine the theoretical bounds of performance.

\addtolength{\textheight}{-12cm}   % This command serves to balance the column lengths
                                  % on the last page of the document manually. It shortens
                                  % the textheight of the last page by a suitable amount.
                                  % This command does not take effect until the next page
                                  % so it should come on the page before the last. Make
                                  % sure that you do not shorten the textheight too much.

%%%%%%%%%%%%%%%%%%%%%%%%%%%%%%%%%%%%%%%%%%%%%%%%%%%%%%%%%%%%%%%%%%%%%%%%%%%%%%%%

%%%%%%%%%%%%%%%%%%%%%%%%%%%%%%%%%%%%%%%%%%%%%%%%%%%%%%%%%%%%%%%%%%%%%%%%%%%%%%%%

%%%%%%%%%%%%%%%%%%%%%%%%%%%%%%%%%%%%%%%%%%%%%%%%%%%%%%%%%%%%%%%%%%%%%%%%%%%%%%%%

% \section*{ACKNOWLEDGMENT}

% The preferred spelling of the word ÒacknowledgmentÓ in America is without an ÒeÓ after the ÒgÓ. Avoid the stilted expression, ÒOne of us (R. B. G.) thanks . . .Ó  Instead, try ÒR. B. G. thanksÓ. Put sponsor acknowledgments in the unnumbered footnote on the first page.

%%%%%%%%%%%%%%%%%%%%%%%%%%%%%%%%%%%%%%%%%%%%%%%%%%%%%%%%%%%%%%%%%%%%%%%%%%%%%%%%

% References are important to the reader; therefore, each citation must be complete and correct. If at all possible, references should be commonly available publications.

\bibliographystyle{IEEEtran}
% \IEEEtriggeratref{9}
\bibliography{reference}

\end{document}